\documentclass[letterpaper]{article} 
\usepackage{aaai2027} 
\usepackage{times} 
\usepackage{helvet} 
\usepackage{courier} 
\usepackage[hyphens]{url} 
\usepackage{graphicx} 
\usepackage{natbib} 
\usepackage{caption} 
\usepackage{amsmath}
\usepackage{amssymb}
\usepackage{array}
\usepackage{booktabs}
\usepackage{subcaption}
\usepackage{listings}
\usepackage{xr}
\graphicspath{{figures/}}

\lstdefinestyle{jsonsmall}{
  basicstyle=\ttfamily\footnotesize,
  columns=fullflexible,
  breaklines=true,
  frame=single,
  framerule=0.3pt,
  keepspaces=true
}

\newcommand{\holo}{Holo-3.1-35B-A3B}
\newcommand{\qwen}{Qwen3.6-35B-A3B}

\title{Why Are GUI Agents Correct but Late? Decode on the Decision-Time
Critical Path, Tested with Pre-Compiled Policy Trees}

\author{
Zihan Dong\textsuperscript{\rm 1},
Rui Qian\textsuperscript{\rm 2},
Qishi Zhan\textsuperscript{\rm 3},
Dongshen Peng\textsuperscript{\rm 4},
Kaixin Li\textsuperscript{\rm 5},
Yu Li\textsuperscript{\rm 6}
}

\affiliations{
\textsuperscript{\rm 1}Georgia Institute of Technology\\
\textsuperscript{\rm 2}Fudan University\\
\textsuperscript{\rm 3}Marquette University\\
\textsuperscript{\rm 4}University of North Carolina at Chapel Hill\\
\textsuperscript{\rm 5}National University of Singapore\\
\textsuperscript{\rm 6}Southeast University, Nanjing, China\\
\texttt{zdong312@gatech.edu}
}

\begin{document}

\maketitle

\begin{abstract}

Computer-use agents often fail on transient GUI events because they produce the correct action only after the relevant window has already closed. We identify the main cause as expensive autoregressive decoding on the decision-time critical path. To this end, we propose Adaptive Anticipatory Policy Trees (AAPT), which eliminates this delay without modifying the underlying model. During idle screen periods, the same frozen multimodal model constructs a bounded conditional policy tree with observable guards, pre-authorized actions, and branch-specific deadlines. The tree is sized to cover the model's own decoding latency. When an event occurs, a lightweight observer matches change-gated frames to a prepared branch and immediately executes the corresponding action without generating new text. In paired trials with pre-registered endpoints and exact McNemar tests, AAPT improves the success rate from 0.50 to 0.79 within a contested decision window ($p=1.8\times10^{-3}$), while producing no incorrect actions. Both open-loop and predict-and-replan baselines achieve zero success because they still decode during execution. A preparation-time sweep shows that the gain emerges where the latency-based tree-sizing rule predicts, and ablations reveal three key requirements: fast observer decoding, valid tree planning, and accurate branch routing. A pre-registered oracle probe rejects our initial hypothesis and instead points to branch routing as the causal bottleneck. We further reproduce the effect on an independent general-purpose multimodal model over 126 paired trials ($p=4.9\times10^{-13}$). On an external benchmark, AAPT matches the overall performance of a reactive baseline, although the two methods exhibit complementary strengths. Together, these results suggest that AAPT performs best when candidate actions can be enumerated in advance, whereas reactive execution remains stronger when they cannot.
\end{abstract}
\section{Introduction}
\label{sec:intro}

Most computer-use agents follow a simple loop: capture a screenshot,
call a multimodal model, execute one action, observe again. This loop
fails when the interface does not wait -- boot prompts,
auto-dismissing dialogs, short-lived authentication requests,
transient toasts, and reaction-time game states all present action
windows that open and close on the environment's clock, not the
model's. The typical failure is not wrong understanding but late
understanding: the agent computes the right action after the window
has already closed.

Two explanations are possible. Either the agent fails to anticipate
the event, or it anticipates correctly but is defeated by where its
computation happens: generating a response with a large multimodal
model takes longer than many of these windows stay open, and calling
such a model at frame rate is not a viable fix. We test which
explanation is correct by holding the model, task, and information
fixed and manipulating only one thing: whether generation occurs on
the decision-time critical path.

\begin{figure}[t]
\centering
\includegraphics[width=1\columnwidth]{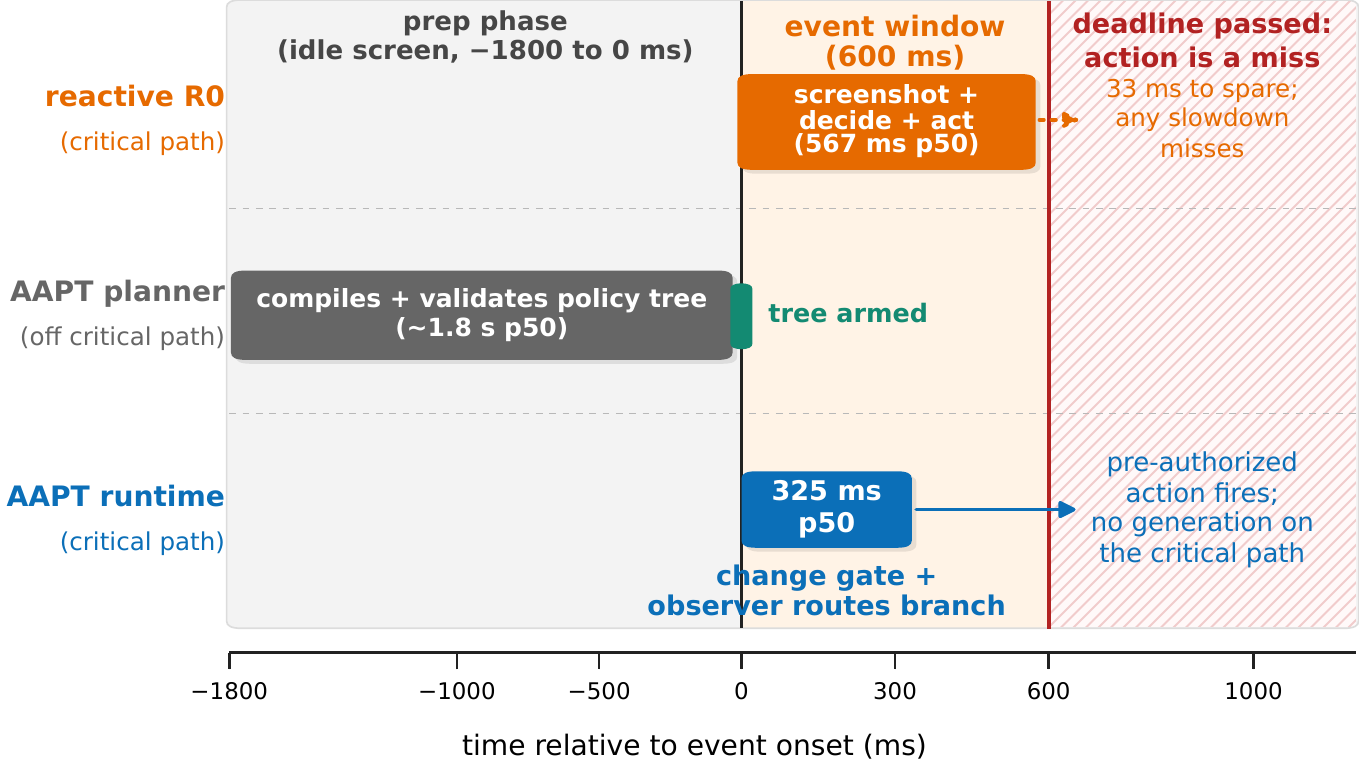}
\caption{The critical path under a contested 600\,ms window (latencies
are measured values). The reactive loop must complete a full
perceive-reason-act round trip (567\,ms p50) inside the window; AAPT
compiles its policy beforehand and took a low-token routing call
($\sim$325\,ms p50) afteward.}
\label{fig:timing}
\end{figure}

We introduce Adaptive Anticipatory Policy Trees (AAPT) to run this
test. While the screen is quiet, the same frozen model precompiles a
small conditional policy tree -- observable guards, pre-authorized
actions, per-branch deadlines -- sized to cover its own decoding
latency. At event time, a lightweight observer under a strict token
budget watches incoming frames and does one cheap thing: identify
which prepared branch matches, and fire its pre-authorized action.
Anything outside the tree halts execution and triggers replanning
(Figure~\ref{fig:timing}). No component of AAPT is individually novel
(Section~\ref{sec:related}); what is new is using it as a controlled
manipulation to isolate a single cause. Every comparison is paired
per-seed on frozen models with exact McNemar tests, and all endpoints
were pre-registered.

This design lets us test three falsifiable predictions: (i) if the
problem were purely a lack of anticipation, acting early while still
decoding on the critical path should already help -- it does not;
(ii) if the problem is decode latency, removing it should help once
preparation time exceeds the planner's own latency, with a predicted
onset; (iii) if raw speed were sufficient, a fast model that cannot
compile or route a valid policy tree should still benefit -- it does
not. We report results for all three, including a pre-registered
prediction of our own that failed, which turns out to strengthen
rather than weaken the causal claim (Section~\ref{sec:results-mechanism}). Our contributions are:
\begin{itemize}
\item A controlled method (AAPT) that isolates decode latency as a
cause of missed action windows, rather than proposing a new agent
architecture per se.
\item Evidence, under pre-registered endpoints and exact paired tests,
that removing decode-time generation -- not earlier action alone --
recovers performance on contested windows, and that this effect
follows the latency-based sizing rule that predicts it.
\item An ablation isolating the capabilities the effect depends on
(fast observer decoding, valid tree planning, accurate routing), and a
replication of the full effect on an independent, untuned model.
\item A boundary result on an external benchmark showing where this
approach helps and where it does not: precompiled trees win when the
correct response is enumerable in advance, and lose otherwise.
\end{itemize}

\section{Related work}
\label{sec:related}

\begin{table}[h]
\centering
\caption{Comparative map of the closest prior systems. $\bullet$ =
implemented and evaluated; ($\bullet$) = partial (qualifier beneath);
$\circ$ = absent.}
\label{tab:comparative}
\footnotesize
\setlength{\tabcolsep}{1.5pt}
\newcommand{\compcell}[2]{%
  \shortstack[c]{#1\\[-1pt] {\scriptsize\shortstack[c]{#2}}}%
}

\begin{tabular}{@{}p{0.80in}
    >{\centering\arraybackslash}p{0.56in}
    >{\centering\arraybackslash}p{0.54in}
    >{\centering\arraybackslash}p{0.57in}
    >{\centering\arraybackslash}p{0.55in}@{}}
\toprule
Work
& \shortstack[c]{Alt.\\actions}
& \shortstack[c]{Live\\routing}
& \shortstack[c]{Model-free\\execution}
& Deadlines \\
\midrule
WebDreamer
& \compcell{($\bullet$)}{candidate\\lookahead}
& $\circ$
& $\circ$
& $\circ$ \\[5pt]

MobileDreamer
& \compcell{($\bullet$)}{prediction tree}
& \compcell{$\circ$}{advises only}
& $\circ$
& $\circ$ \\[5pt]

TraceR1
& \compcell{$\circ$}{one trajectory}
& \compcell{$\circ$}{replans\\each step}
& $\circ$
& $\circ$ \\[5pt]

Speculative Actions
& \compcell{($\bullet$)}{top-$k$\\breadth}
& \compcell{($\bullet$)}{slow-actor\\commit}
& \compcell{($\bullet$)}{speculative\\fast path}
& \compcell{($\bullet$)}{API latency} \\[5pt]

AOI
& $\circ$
& \compcell{($\bullet$)}{captures,\\no routing}
& $\circ$
& \compcell{($\bullet$)}{perception\\only} \\[5pt]

PreAct
& \compcell{($\bullet$)}{from\\experience}
& \compcell{$\bullet$}{verify-then-\\replay}
& $\bullet$
& \compcell{($\bullet$)}{repeat\\efficiency} \\[3pt]

\midrule
AAPT
& \compcell{$\bullet$}{contingency\\tree}
& \compcell{$\bullet$}{guard routing}
& \compcell{$\bullet$}{within tree}
& \compcell{$\bullet$}{coverage +\\deadlines} \\
\bottomrule
\end{tabular}
\end{table}

\paragraph{GUI world models and anticipatory planning.}
Recently, a number of GUI world models have emerged that predict
future UI states to score or search over candidate actions before
execution~\citep{gu2024webdreamer,luo2025vimo,guan2026cuwm,cao2026mobiledreamer,zheng2026code2world,xiao2026webworld,ding2026dynaweb},
including MobileDreamer's recursive tree-of-prediction over candidate
actions and states. In all of these, the imagined future is
deliberative context consumed once at decision time: no observer
routes live frames through a prepared tree, and no horizon is set by
the planner's own decoding latency. Building on this predictive
capability, a separate line acts on unconfirmed futures directly.
Receding-horizon methods such as TraceR1~\citep{liang2026tracer1}
predict a trajectory, execute its first action, and replan --
keeping one expensive model call on every step's critical path, a
pattern our P1 baseline instantiates and which scores zero in the
contested regime -- while StreamAgent~\citep{yang2025streamagent}
anticipates future evidence for perception rather than action
commitment. A parallel line hides latency through speculation instead:
Interactive Speculative Planning~\citep{hua2024interactive}, Dynamic
Speculative Planning~\citep{guan2025dsp}, Speculative
Actions~\citep{ye2025speculativeactions}, and
AgenticCache~\citep{kim2026agenticcache} overlap fast approximate
execution with slow verification, or replay cached plan transitions
under asynchronous checks, all resting on speculated work being
reversible. AAPT inverts this: nothing executes speculatively; several
actions are merely prepared, and one fires only once a live
observation satisfies its guard within deadline -- the discipline GUIs
require, since plausible actions are not safely pre-launchable (See more in Table~\ref{tab:comparative}). 

\paragraph{Continuous and dynamic GUI observation.}
The Agent-Computer Observation Interface~\citep{li2026aoi} continuously
samples the screen and gates changed frames, and
DynamicGUIBench~\citep{liu2026dynamicgui} and living-screen
benchmarks~\citep{yao2026livingscreen} formalize continuous-time GUIs
where observation itself bears cost. This line improves what the agent
sees, not how fast it can act on what it sees. AAPT adds the
conditional-execution layer above this observation layer; our transfer
evaluation runs on the DynaCU-Bench tasks introduced alongside AOI.

\paragraph{Verified replay and procedural memory.}
A related family -- PreAct~\citep{li2026preact},
SkillDroid~\citep{chen2026skilldroid},
ActionEngine~\citep{zhong2026actionengine}, Agent Workflow
Memory~\citep{wang2024awm}, and SKILL.nb~\citep{elhattami2026skillnb}
-- compiles successful past trajectories into verify-before-replay
policies or reusable gated skills, which requires the trajectory to
have already been experienced. AAPT instead synthesizes a short-lived
contingency policy for unseen futures; promoting
validated AAPT subtrees into this kind of long-lived memory is
complementary future work. More broadly, conditional branches over
action-observation pairs are themselves the native representation of
POMDP contingency
planning~\citep{ye2017despot,meuleau2003olcp,blumenthal2023contingent,he2011macro}
and behavior trees~\citep{iovino2020behaviortrees}; 
\begin{figure}[h]
\centering
\includegraphics[width=0.98\columnwidth]{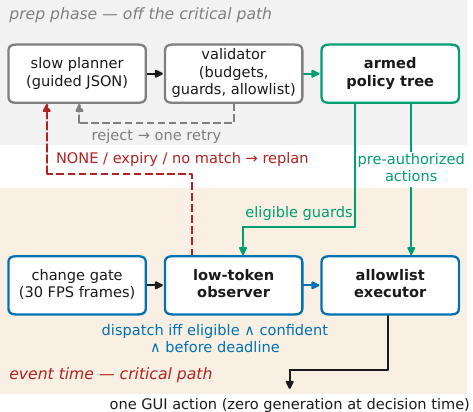}
\caption{AAPT dataflow. Off the critical path, the slow planner compiles
and validates a policy tree during the prep phase. On the critical path,
event-time execution is a change gate, one low-token observer call, and a
pre-authorized lookup -- no generation. Unmatched or expired branches
fall back to replanning.}
\label{fig:dataflow}
\end{figure}

\section{Adaptive Anticipatory Policy Trees}
\label{sec:method}

\subsection{Setting, success criterion, and the latency-coverage rule}
\label{sec:method-horizon}
We consider environments with transient, sometimes irreversible decision
points, where success couples correctness with timing: $\text{success} =
\text{correct action} \wedge \text{action before deadline}$. AAPT has the
slow model prepare, in advance, enough conditional behavior to keep the
system operational through the interval when the next slow reasoning
call is unavailable. This isolates generation on the decision-time critical path while keeping the model, information, and task fixed. (Figure~\ref{fig:dataflow}), so the presence or absence of the advantage separates the causal accounts of Section~\ref{sec:intro}.

This intuition has a quantitative form. Let $L_{p95}$ be the
95th-percentile end-to-end planner latency, $M$ a safety margin
($500$\,ms in all experiments), and $T_{\mathrm{cover}}$ the wall-clock
interval a prepared tree stays valid for. The tree must bridge the
planner's own unavailability:
\begin{equation}
T_{\mathrm{cover}} \;\geq\; L_{p95} + M .
\label{eq:cover}
\end{equation}
Eq.~\eqref{eq:cover} is the hypothesis's quantitative content, not a
tuning heuristic: from independently measured quantities, it names the
preparation budget below which the advantage cannot exist and above
which it appears, a prediction the prep-budget sweep tests directly --
the measured crossover brackets $L_{p95}+M \approx 2.4$\,s. With
$\mathbb{E}[\Delta t]$ the expected transition duration, the horizon
$n \approx \lceil T_{\mathrm{cover}} / \mathbb{E}[\Delta t] \rceil$ is
likewise a latency-coverage quantity, not an open-ended lookahead depth.
Trees whose declared coverage misses the measured budget (3{,}500\,ms
here) are rejected at validation.

\subsection{Tree representation}
\label{sec:method-tree}
A policy node is a tuple $v=(z_v, C_v, A_v, D_v, Q_v, R_v, F_v)$: a
predicted state description, an observable guard, a pre-authorized
primitive action or short chunk, an action deadline, a calibrated
confidence, a risk class, and fallback behavior. The fallback is realized
as the runtime abstain path (Section~\ref{sec:method-runtime}), not as a
tree node. The tree is flat rather than recursive: each node's guard is
evaluated independently against the current observation, so routing
reduces to a single classification over sibling guards rather than a
multi-step traversal, and weaker schema-followers reliably fail to emit
correct recursive structures under this same time budget
(Section~\ref{sec:results-boundary}). Implementation details -- the
JSON schema, a canonical compiled example, and the executor's
action-allowlist enforcement -- are given in Appendix~\ref{app:schema}.

\subsection{Adaptive branching under uncertainty}
\label{sec:method-branching}
At generation time, the planner assigns each node a branching factor
$0 \leq K_v \leq K_{\max}$ following a three-way rule:
\begin{equation}
K_v =
\begin{cases}
1, & \text{confidence} \geq \tau_c \text{ and risk} \leq \tau_r,\\
\min(K_{\max}, m_v), & \text{outcome uncertain},\\
0, & \text{safe fallback / replan required},
\end{cases}
\label{eq:branching-rule}
\end{equation}
i.e. a node commits to a single successor when the action is both
confident and low-risk, opens $m_v$ parallel branches when the action
is certain but its outcome is not, and generates no child -- routing to
the runtime abstain path (Section~\ref{sec:method-runtime}) -- when
neither condition holds. $K_v=0$ here is consistent with
Section~\ref{sec:method-tree}: the fallback is realized as the runtime
abstain path, not as a tree node.

For the uncertain-outcome case, $m_v$ is the smallest branch count that
jointly covers the likely outcomes:
\begin{equation}
\begin{aligned}
m_v &= \min\Big\{K : \sum_{i=1}^{K} P(s_{v,i}) \geq \rho\Big\}, \\
K_v &= \min(K_{\max}, m_v).
\end{aligned}
\label{eq:coverage-target}
\end{equation}
Probability mass not represented by explicit branches is owned by the
abstain path rather than by an armed default branch
(Section~\ref{sec:method-tree}).

Two different enforcement mechanisms apply to
Eq.~\eqref{eq:branching-rule}--\eqref{eq:coverage-target}. The
structural budgets -- $K_{\max}$, maximum depth $n_{\max}$, total node
count, the coverage-time budget $T_{\mathrm{cover}}$
(\texttt{cover\_ms}), and risk-specific open-loop horizons -- are hard
limits, checked numerically by the post-hoc validator
(Appendix~\ref{app:schema}). The decision thresholds $\tau_c$,
$\tau_r$, $\rho$, and the branch count $m_v$ are not runtime-enforced;
they are design rules realized as planner-prompt instructions that ask
the planner to branch on uncertain outcomes and cover the likely ones,
so their satisfaction depends on the planner's compliance with the
prompt rather than on any executor-side check.

\subsection{Runtime: observe, route, execute}
\label{sec:method-runtime}

The fast path never generates free-form actions. The screen is captured at a fixed frame rate, and a pixel-difference change gate suppresses frames in which the scene has not changed; a periodic probe frame is forced through independently of the gate so that a stalled screen cannot silence the observer indefinitely (configuration values given in Section~\ref{sec:setup}).

On a changed frame the low-token observer receives the frame and only the currently eligible branch guards, and returns a single JSON object \texttt{\{branch\_id, confidence, reason\}}: a short, fixed-schema decode at low temperature, chosen so the routing decision is deterministic given the input, with \texttt{NONE} as the abstain value.

The runtime applies three suppression checks: the returned id is currently eligible; the planner's stored confidence for the branch clears a fixed threshold $\tau_m$; and the decision age does not exceed the branch deadline. The confidence check has a sharp boundary. The planner writes the explicit outcome probabilities into $Q_v$, If the largest prior for a task family is below $\tau_m$, that family becomes unroutable (Appendix~\ref{app:round1-expA}).

The action must also be on the pre-authorized allowlist; after firing, only the node's children become eligible; a no-match, ambiguity, expiry, or verification failure falls back to replanning. By construction, a branch whose routing decision arrives after its deadline is suppressed rather than executed -- Section~\ref{sec:results} reports the consequences of this guarantee for action correctness.
\section{Experimental setup}
\label{sec:setup}

\begin{figure}[t]
\centering
\includegraphics[width=1\columnwidth]{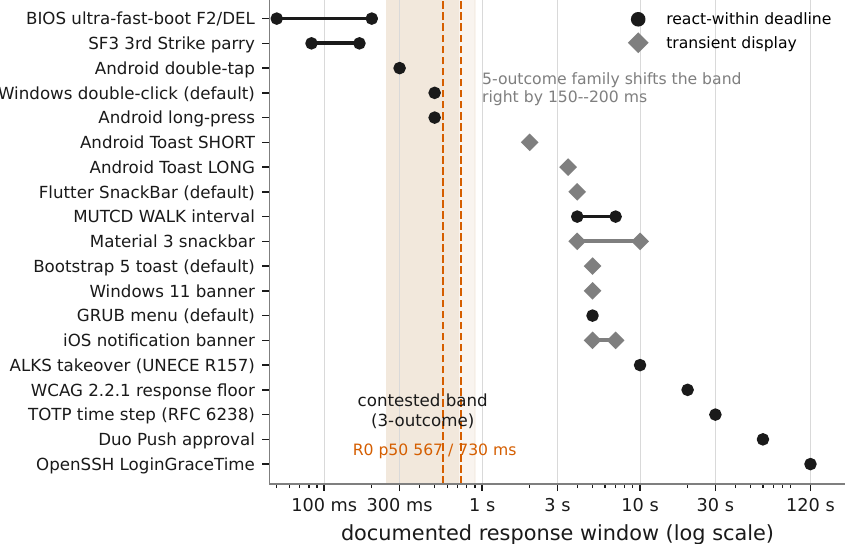}
\caption{Documented real-world response windows (log scale) against the benchmark's contested band (shaded; the five-outcome family shifts it
right by 150--200\,ms) and the measured reactive p50 (dashed). Below the band a reactive loop is structurally excluded; inside it, contested; an order of magnitude above it, sufficient.}
\label{fig:deadline-survey}
\end{figure}

\subsection{The contested-window benchmark}
\label{sec:setup-benchmark}
Existing GUI-agent benchmarks such as OSWorld, WebArena, Mind2Web, and AndroidWorld~\citep{xie2024osworld, zhou2023webarena, deng2023mind2web, rawles2024androidworld} do not provide the millisecond-scale control required to isolate deadline-sensitive action selection. We therefore built \texttt{key\_prompt}, a timed-GUI benchmark with three properties: adjustable 250--2000\,ms event windows for locating the contested regime, deterministic server-side scoring without a judge model, and strict control over all pre-event information.
This range is grounded in real deployed systems rather than chosen arbitrarily. Figure~\ref{fig:deadline-survey} compares it with 19 documented response windows (Appendix~\ref{app:round1-expE}), which fall into three regimes relative to the measured reactive p50 of 567--730,ms.
After a trial is armed, a transient prompt appears following a fixed 750\,ms delay and requests input. The requested key is determined by a hidden, seed-derived schedule, and the prompt disappears when the event window expires. A trial succeeds only if the requested keypress reaches the server before the deadline; a different key is recorded separately as an incorrect action. The agent is told only that a temporary prompt will request one of the three keys. Neither the delay value, a countdown, nor any branch-specific cue is visible before the event. Consistent with this design, R1, which can exploit timing but cannot observe the requested key, scores zero, while randomizing the delay over $[500,1000]$\,ms leaves T2's advantage unchanged (discordant $15/0$, $p=6.1\times10^{-5}$; Appendix~\ref{app:round1-expB}). Every comparison uses a paired per-seed design. For each seed, all conditions receive the byte-identical hidden schedule, with condition order alternated, and differences are assessed using the exact two-sided McNemar test. Seeds are drawn from pre-registered, disjoint, key-balanced ranges and are never reused (Appendix~\ref{app:protocol}). Because reaction latency is nearly deterministic, absolute success rates at a fixed window can vary sharply across seed draws, whereas within-seed comparisons remain valid. The contested window, defined as the region where the reactive baseline is neither at floor nor ceiling, is therefore recalibrated for each stage using reactive-only scans conducted before treatment runs; all deviations from pre-registered windows are reported in Section~\ref{sec:results}. All experiments use six frozen checkpoints, with the same checkpoint serving as planner and observer within each AAPT run (Table~\ref{tab:crossmodel}; Appendix~\ref{app:models}, Table~\ref{tab:models}). A fixed pixel-difference change gate runs at 30\,FPS, adding at most one frame interval ($\leq 33$\,ms) between event onset and the observer's input.

\subsection{Conditions}
\label{sec:setup-conditions}

To test whether removing full-model decoding from the event-time critical
path improves deadline-limited performance, we compare AAPT with four
baseline and diagnostic conditions. We define \textbf{R0} as the reactive
baseline: after each screenshot, the model generates one action, so every
response requires a full model round trip after the event appears.
\textbf{T2} is our AAPT condition: before the event, a slow planner
compiles a bounded policy tree; after the change gate fires, a low-token
observer selects a branch and executes its pre-authorized action subject
to eligibility, confidence-floor, and deadline checks. \textbf{R1}, the
open-loop baseline, prepares a short unconditional action sequence before
the event and executes it without observing which prompt appears.
\textbf{P1}, predict-and-replan, predicts two steps, executes one, and
replans after each observation, providing a receding-horizon baseline in
the spirit of TraceR1~\citep{liang2026tracer1}. \textbf{O1}, oracle
routing, uses ground truth to select the correct branch of the compiled
tree at zero decision latency while retaining the same change gate. R1
and P1 test whether acting early alone is sufficient, whereas O1
estimates the performance lost to observer decoding and routing errors.

\subsection{Serving invariants, reproducibility, and pre-registration}
\label{sec:setup-serving}

Both AAPT roles decode under a JSON-Schema grammar via vLLM's \texttt{guided\_json}, backed by xgrammar~\citep{dong2024xgrammar}, with the planner at temperature 0.2 and a scalar-only action schema; a tree
that fails post-hoc validation triggers one bounded resample at unchanged
temperature. Schema-following on this stack is knife-edge sensitive to
kernel-level numerics, so the \texttt{torch.compile} cache key is treated
as part of the experimental condition: it is verified before every trial,
frozen per model in provenance, and any change voids collected data. No
fine-tuning was used anywhere in this pipeline (validity numbers and both
engineering incidents are reported in Appendix~\ref{app:provenance} and
Appendix~\ref{app:method-details}). Each stage was also specified in a
written plan before execution, including windows, seed ranges, primary
endpoints, decision rules, and acceptance criteria, with deviations
reported alongside the results they affect. The strongest instance is the
gate confirmation in Section~\ref{sec:results-replication}: the two
confirmation models, their predicted gate outcomes, and the interpretation
of every possible result were committed to the repository before either
model produced a single number. No model was added after results were seen,
and no seed re-rolls or post-hoc sample-size increases occurred.
\begin{figure*}[h]
\centering
\includegraphics[width=0.95\textwidth]{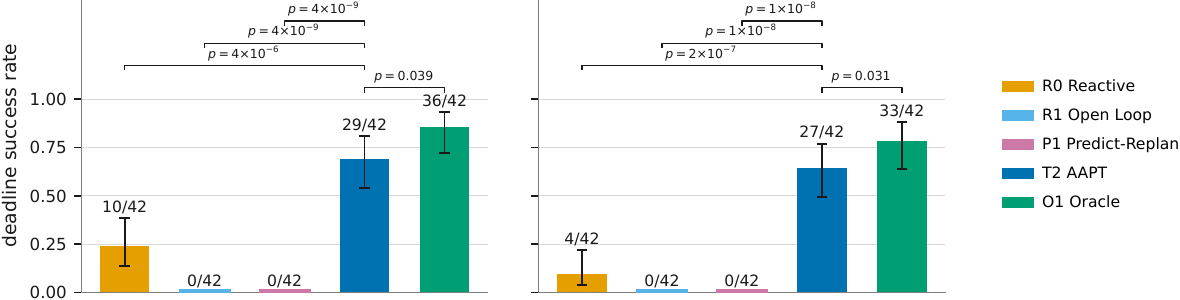}
\caption{Mechanism isolation at the two contested windows (left: 650\,ms; right: 600\,ms; 42 pairs each; Wilson 95\% intervals; brackets give exact McNemar $p$ for T2 against each condition). R1 and P1 also act before a full round trip yet score zero, drawn as outlined stubs so 0/42 is a visible measurement, while O1 bounds what an error-free observer could extract from those trees.}
\label{fig:stagec}
\end{figure*}

\begin{table*}[t]
\centering
\caption{Cross-model gate matrix: the gain appears where all
three gates (column heads) pass; ``n/r'' = not reached. $^1$~After one documented prompt-adaptation round. $^2$~A
ground-truth-routed probe (O1) flips this to 22/42 vs.\ 12/42, $p=0.021$
(Appendix~\ref{app:round1-expD}). $^3$~Stopped by protocol: no branch set
to route (\S\ref{sec:results-replication}).}
\label{tab:crossmodel}
\footnotesize
\setlength{\tabcolsep}{4.5pt}
\begin{tabular}{lcccl}
\toprule
Model & (a) decode $<1$\,s & (b) flat-tree validity & (c) branch acc.\ $\geq 0.85$ & Outcome (T2 vs.\ R0) \\
\midrule
Holo-3.1-35B-A3B (MoE) & \checkmark~$\sim$0.2\,s & \checkmark~$\sim$1.0 & \checkmark~0.86--0.93 & T2$\gg$R0 ($\times$4 blocks) \\
Qwen3.6-35B-A3B (MoE) & \checkmark~0.15--0.45\,s & \checkmark~1.00 & \checkmark~0.969 & T2$\gg$R0 ($\times$3 blocks) \\
UI-Venus-1.5-30B-A3B (MoE) & \checkmark~$\sim$0.15\,s & \checkmark~0.95$^1$ & $\times$~0.39 & n.s.\ (11/42 vs.\ 13/42)$^2$ \\
EvoCUA-32B (dense) & $\times$~$\sim$2.5\,s & \checkmark~0.98 & n/r & T2 0/43 vs.\ R0 22/43 \\
UI-Ins-32B (dense) & $\times$~$\sim$1.5\,s & $\times$~0.10$^1$ & n/r & T2 0/50 (R0 $\approx 0.5$) \\
UI-TARS-1.5-7B (dense) & \checkmark~0.5--1.1\,s & $\times$~0.50$^1$, cover 0/30 & n/r & paired block not run$^3$ \\
\bottomrule
\end{tabular}
\end{table*}

\section{Results}
\label{sec:results}

Unless stated otherwise, all results use frozen \holo{} on \texttt{key\_prompt}, with 42 per-seed paired trials per window analyzed
by exact two-sided McNemar tests; serving invariants were verified before
each block, and the complete pre-registered prediction ledger appears in
Appendix~\ref{app:ledger-r1}.

\subsection{Test 1: the contested-window advantage}
\label{sec:results-headline}

\paragraph{Below the reactive floor.}
At 500\,ms, below R0's reaction floor, T2 succeeds 28/42 against R0
0/42, all 28 discordants favoring T2 ($p=7.5\times10^{-9}$); at relaxed
windows both reach ceiling and at 250\,ms both floor. The stronger claim
is that T2 beats a functioning reactive baseline. A four-window scan
(Appendix~\ref{app:scan}) locates the contested regime at 600--650\,ms,
where T2 beats a degraded-but-alive R0 (Figure~\ref{fig:stagec}). We
report a deviation: the pre-registered primary was 700\,ms, where R0
turned out near ceiling and not significant ($p=0.25$), and we do not
claim it; this stage rests on the pre-registered secondary windows.
Incorrect actions were 0\% for both conditions at every window.

\paragraph{Declared-primary confirmation.}
A confirmatory re-run closed that deviation, with 650\,ms declared as
primary before data collection, on the tuned configuration and fresh
seeds. The primary passes: T2 33/42 (0.79) against R0 21/42 (0.50),
$p=1.8\times10^{-3}$, with every secondary window also significant,
incorrect-action 0 everywhere, tree validity 1.0, and observer branch
accuracy 0.86--0.93. This is the headline result, a pre-registered,
declared-primary win over a live reactive baseline in the contested
regime. The zero incorrect-action rate is a property of the execution
design at this budget setting -- a late or unmatched decision is
suppressed rather than executed (Section~\ref{sec:method-runtime}) --
not a general empirical safety guarantee: Appendix~\ref{app:round1-expA}
shows wrong keys fire once the branch budget is mis-set.

\paragraph{Costs and misses.}
The win is not free. T2 generates $3.2\times$ R0's completion tokens per
trial, almost all of it tree compilation off the critical path (283.5
tokens, p50 1.81\,s); the runtime observer is cheaper per call than R0's
decisive calls (59.3 against 105.8 tokens; full ledger in
Appendix~\ref{app:cost}). Of T2's nine misses on this block, five are
correct routes converted too late, three are wrong-branch routes whose
presses also landed post-deadline, and one is a correctly suppressed late
decision. Late conversion, not tree quality, dominates, which agrees with
the O1 headroom analysis below.

\subsection{Test 2: pre-computation is the mechanism}
\label{sec:results-mechanism}

Figure~\ref{fig:stagec} compares all five conditions at the contested
windows; no condition produced an incorrect action. R1 and P1 fail for a
measured critical-path reason. Each reactive-family condition needs two
model calls (a \texttt{WAIT}, then the decisive action), and anticipatory
prompting lengthens the second output: measured per-decision paths are
273\,ms/30 tokens for the T2 observer, against 411\,ms/48 for R0,
459\,ms/60 for R1, and 484\,ms/66 for P1. Acting early therefore does not
help by itself; anticipation becomes a latency win only when its reasoning
moves off the critical path. O1 sits above T2 at both windows, but neither comparison survives Holm--Bonferroni, so the gaps are estimated headroom rather than confirmed effects. The direction is consistent (discordants 8/1 and 6/0), and all eight 650\,ms oracle-only pairs are \texttt{routed\_late} misses for T2, never wrong branches. The binding constraint is therefore the observer, not the planner.

\subsection{Test 3: replication within the declared class}
\label{sec:results-replication}

\paragraph{Fresh-seed repeats (same model).}
Two further 42-pair blocks on fresh seeds at a 550\,ms sub-cliff window
replicate the direction: pooled 84 pairs, T2 0.619 against R0 0.143, 40
T2-only pairs and zero R0-only ($p=1.8\times10^{-12}$). These runs also
exposed the knife-edge property motivating the paired design
(Section~\ref{sec:setup-benchmark}): the shift is per-seed-set difficulty
variance, not time drift (Appendix~\ref{app:window-sensitivity}).

\paragraph{Pre-registered confirmation on an untuned generalist.}
The boundary map (Section~\ref{sec:results-boundary}) yields a falsifiable
hypothesis: the gain appears when a model has (a) sub-second observer
decode, (b) schema-valid flat-tree planning under guided decoding, and (c)
branch-routing accuracy $\geq$0.85. We tested this confirmatorily. Two
models were selected by predicted gate outcome, with predictions and their
interpretation committed before any data. \qwen{}, an untuned generalist
of Holo's architecture class, passed all three gates
(Table~\ref{tab:crossmodel}), and the pre-declared consequence followed:
at the declared 600\,ms primary, the pooled 126 pairs yield T2 0.778
against R0 0.341, with 60/5 discordants ($p=4.9\times10^{-13}$) and every
block individually significant. Thus the gain is neither Holo-specific
nor dependent on an agentic fine-tune. The pre-fixed claim ceiling still
applies: this upgrades the scope to the Qwen3.5-MoE class, including an
untuned generalist, but the shared architecture family does not establish
architecture-agnostic transfer.

The second pre-registered model, UI-TARS-1.5-7B, was predicted at risk on
gate (b) and failed before a paired block could be run
(Table~\ref{tab:crossmodel}). This is a contract-expression boundary, not
a claim about the model's GUI skill.

\subsection{Test 4: three capability gates; routing is causal}
\label{sec:results-boundary}

Table~\ref{tab:crossmodel} summarizes all six models. T2$>$R0 appears on
the two models that pass all three gates, and each failing model localizes
a different gate. The routing gate is made causal by a pre-registered
oracle probe on UI-Venus-1.5-30B-A3B, which is fast and schema-valid but
routes wrongly (branch accuracy 0.39) and its paired block is a tie
($p=0.625$). Ground-truth routing over the same frozen planner flips the
tie to a win (22/42 against 12/42, $p=0.021$), against our own committed
prediction, with success equal to \{schema-valid tree $\wedge$ true
outcome covered\}. The probe session's trees had higher coverage than the
original block's (recall 0.59 against 0.22), so the licensed statement is
conditional: when tree coverage clears the reactive baseline, routing
quality alone separates tie from win, and gate~(c) and planner
outcome-recall are jointly binding (Appendix~\ref{app:round1-expD}).

\subsection{Test 5: transfer requires pre-enumerability}
\label{sec:results-transfer}

To test transfer beyond the local benchmark, we paired frozen reactive
and AAPT agents on 39 deterministic, deadline-focused DynaCU-Bench
tasks~\citep{li2026aoi}. A pre-registered majority-of-3 run produced an
aggregate tie: reactive 7/39 versus AAPT 6/39, with 5/4 discordants
($p=1.0$). The wins nevertheless separate cleanly by task structure
(Appendix~\ref{app:dynacu}): all four AAPT-only wins are dashboard
deadlines whose responses are pre-enumerable at compile time, where
pre-armed actions were dispatched 140--215\,ms after the gated frame
while the reactive agent's $\sim$437\,ms round trip missed the window.
All five reactive-only wins require multi-step sequencing, late-revealed
values, or positional grounding unavailable at compile time. Thus,
pre-computation transfers its deadline advantage only when both outcomes
and actions can be enumerated in advance; it complements rather than
replaces reactive execution.
\subsection{Test 6: the sizing rules hold quantitatively}
\label{sec:ablation}
\label{sec:ablation-kn}
\label{sec:ablation-observer}
\label{sec:ablation-window}

\begin{figure*}[t]
\centering
\includegraphics[width=0.85\textwidth]{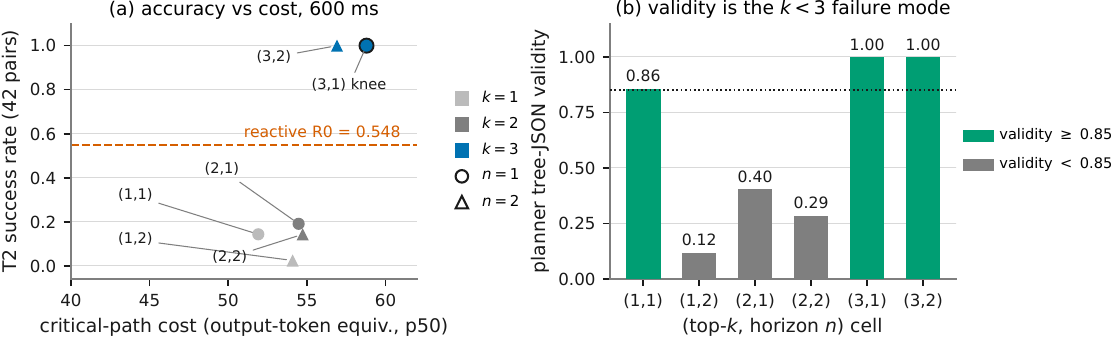}
\caption{Budget frontier at the 600\,ms primary window (42 pairs per
cell, shared reactive arm). (a) T2 success versus critical-path cost
($k$ = color, $n$ = marker): the knee $(k{=}3,n{=}1)$ is the smallest
budget covering all three outcomes. Costs vary by less than 15\% across
cells. (b) Planner tree-JSON validity against the 0.85 gate: budgets below
the outcome-set size fail mainly through low validity, with $k{=}2$ worst.}
\label{fig:frontier}
\end{figure*}

These experiments test the sizing predictions of
Sections~\ref{sec:method-horizon} and~\ref{sec:method-branching}. In the
$k\times n$ sweep at the 600\,ms primary window, performance changes
sharply once the branch budget matches the three outcomes: the smallest
sufficient configuration, $(k{=}3,n{=}1)$, succeeds on all 42 pairs and
beats the reactive arm with 19/0 discordants
($p=3.8\times10^{-6}$), while $n=2$ provides no benefit.
Underbudgeted cells are not meaningfully cheaper—their critical-path cost
varies by less than 15\%—but fail because planner validity falls below
the 0.85 gate. A five-outcome replication shifts the required budget with
the outcome-set size and shows that mis-set budgets can cause wrong
actions, not only misses (Appendix~\ref{app:round1-expA}). Although
$(3,2)$ permits a deeper tree, the planner constructs the same four-node
depth-1 policy as at $(3,1)$, making $(3,1)$ the recommended operating
point. The observer sweep selects a 64-token cap, and window recalibration
tracks seed-set difficulty rather than server drift
(Appendix~\ref{app:window-sensitivity}).

\section{Discussion}
\label{sec:discussion}

\paragraph{Anticipation is a pipeline, not a model trait.}
The cross-model results replace the vague question of whether a model can
``anticipate'' with a factorized systems criterion. In our implementation,
the AAPT success event is contained in
$\{\text{valid tree}\}\cap\{\text{true outcome covered}\}\cap
\{\text{correct route}\}\cap\{\text{route before deadline}\}$.
The three capability gates provide inexpensive probes of this chain:
flat-tree validity tests whether a policy can be represented, branch
accuracy whether it can be retrieved, and observer latency whether that
retrieval remains useful. The UI-Venus oracle probe shows that syntactic validity is not semantic coverage: even perfect routing recovers only outcomes placed in the tree, so performance is set by the weakest planning--acting interface, not by parameter count, GUI fine-tuning, or decode speed alone. This explains why an untuned generalist that passes the pre-declared gates reproduces the gain, while faster or GUI-specialized models that fail one interface do not, and turns a negative result into a component-level diagnosis. The benefit is also a latency--compute trade rather than a universal speedup:
AAPT moves 283.5 planner tokens off the critical path but raises total
completion tokens by $3.2\times$. It is economical only when the reactive
latency distribution overlaps the task deadline and the available quiet
time exceeds Eq.~\ref{eq:cover}; outside that band, pre-computation is
either too late or unnecessary. Within it, the observer is the next lever:
the estimated oracle gap
($\mathrm{O1}-\mathrm{T2}\approx0.14$--$0.17$) indicates that the compiled
trees contain more value than live routing extracts, suggesting that a
distilled guard matcher or OCR/template routes for easy branches could
close part of the gap without increasing planning cost.

\paragraph{When not to use AAPT.}
AAPT trades adaptivity for immediacy. It applies when guards and candidate
actions can be bound before the event; multi-step sequencing,
late-revealed values, and late-revealed coordinates violate that contract
(Section~\ref{sec:results-transfer}). The disjoint AAPT and reactive wins
on DynaCU therefore suggest a hybrid controller whose first decision is
whether the next transition is pre-enumerable, using AAPT as a
latency-specialized path rather than a replacement for reactive execution.

\paragraph{Reproducibility requires preserving the decision boundary.}
This study exposes two hidden experimental variables. First, changing the
compiled-kernel identity moved tree validity from 88.1\% to 18.3\% under
identical weights, prompts, sampler settings, and screenshots by flipping
one schema token. Second, a roughly 100\,ms latency shift can move a fixed
window from contested to ceiling. These are not generic engineering
nuisances: they directly change the validity and deadline terms in the
factorization above. Reproducing a thresholded agent result therefore
requires both functional invariance (model revision, byte-exact launch,
compiled-kernel identity, and component gate metrics) and regime
invariance (latency quantiles and a baseline-only check that the window is
still contested). Without these quantities, mechanism and serving-regime changes are indistinguishable.

\paragraph{Limitations.}
\label{sec:limitations}
The evidence establishes a mechanism, not broad superiority of AAPT. The
positive end-to-end result is confined to two Qwen3.5-MoE checkpoints on
one controlled scenario family; DynaCU is an aggregate tie whose
dissociation rests on 4 versus 5 discordant tasks. Contested-window rates
are seed- and latency-sensitive, so only paired comparisons are
interpretable, and UI-Venus shows that tree coverage can vary even under
frozen recorded settings. Late-bound parameters, positional grounding,
and reactive multi-step behavior remain outside AAPT's current contract;
asynchronous refresh, policy memory, and fine-tuning are untested. The full register appears in Appendix~\ref{app:limitations}.


\section{Conclusion}
\label{sec:conclusion}

Missed transient GUI events can be a scheduling failure rather than a
reasoning failure. AAPT shows that moving generation off the
decision-time critical path can recover otherwise-late actions, provided
the event is covered by the compiled tree and routed before its deadline.
Anticipation is therefore a pipeline property, not a model trait. The
next step is a hybrid controller that detects whether a transition is
pre-enumerable: use AAPT for those cases and reactive execution for
late-bound or multi-step interactions. Distilled routing, asynchronous
tree refresh, and reusable policy memory could reduce latency and
amortize preparation cost.

\newpage
\bibliography{refs}
\section{Protocol details}
\label{app:protocol}

\paragraph{Stage ladder.}
The study ran as a pre-registered ladder of stages, each with a written plan
(goals, windows, seeds, endpoints, decision rules) committed before
execution: pilot and planner diagnostics; guided-decoding adoption and the
below-floor hypothesis test; the contested-window scan; hardening (planner
retry, observer token sweep); the declared-primary confirmation; mechanism
isolation (R1/P1/O1); external transfer; the $k \times n$ ablation; and
replication (fresh-seed repeats, the four-model boundary map, and the
two-model pre-registered gate confirmation).

\paragraph{Windows.}
Relaxed and floor anchors: 250, 500, 1000, 2000\,ms. Contested-regime
windows by stage: scan 600/650/700/750\,ms (700 pre-registered primary,
n.s.); confirmation 650\,ms declared primary with 600/625/675 secondaries;
mechanism isolation 650/600\,ms; ablation 600\,ms primary (recalibrated
from 650 by the reactive-only protocol) with 575/650 secondaries;
fresh-seed repeats 550\,ms (sub-cliff); cross-model blocks at each model's
own recalibrated knee (600\,ms for UI-Venus and the confirmation
generalist; 4--10\,s scans for the slow dense models).

\paragraph{Paired-block anatomy.}
42 pairs per block, key-balanced (14 per outcome), condition order
alternating per pair, identical seed-derived hidden schedule within a pair,
resume-safe manifests, zero tolerated pair-integrity failures. Success,
incorrect-action, failure stage, and all latencies are recorded per trial;
exact two-sided McNemar on discordant pairs is the only significance test
used.

\paragraph{Depth rule for replication blocks.}
Fixed in advance: if a model's primary paired block shows the positive
direction, run two repeat blocks (42 pairs each) at the same window; a null
or negative primary gets one block plus a mechanism read. Boundary models
that fail a capability gate persistently are stopped at the gate and
reported (a paired block against a structurally failing arm can only
reproduce 0/42).

\paragraph{Prompt-adaptation policy.}
Per model, at most one documented round of minimal prompt adaptation is
permitted to meet the planner-validity gate (for grounding-specialized
models this is typically a hardened flat-children instruction); the
adapted prompt is diffed, frozen, and used for all subsequent blocks of
that model. Base-prompt behavior is always reported alongside. The planner
retry-on-invalid is bounded (one resample at unchanged temperature) and its
firing is recorded per trial.

\paragraph{Seed ledger.}
Seeds are integers whose hidden schedules derive deterministically; every
consumed range is burned and never reused. Burned enclosing ranges:
14001--14005 and 40001--40020 (pilot), 48001--48006, 49001--49003,
50001--50062, 51001--51241 (diagnostics), 52001--52075 (guided-decoding
arms), 53001--56047 (window sweep), 57001--60047 (contested scan),
61001--61034 and 62001--62382 (hardening), 63001--63202 and 64001--64018
(confirmation and click-target probe), 65001--65099 (mechanism isolation),
66001--66112 and 66901--66903 (ablation), 67001--67084 (fresh-seed
repeats), 68004--68019, 68100--68161, 68300--68314 (boundary models),
68400--68591 and 68900--68939 (gate confirmation). The external benchmark
uses its own fixed task set (39 tasks, upstream commit pinned) and burns no
local seeds.

\section{Policy-tree JSON Schema}
\label{app:schema}

The guided-decoding schema for the planner (the $k \times n$ ablation
generates per-cell variants of the same shape with adjusted bounds). Note
the flat node list with string-id \texttt{children}; the recursion-free
shape is itself load-bearing (Section~\ref{sec:results-boundary}). For
the three-outcome benchmark event, the canonical compiled tree this
schema yields is:
\begin{lstlisting}[style=jsonsmall]
root: idle screen -> WAIT
 |- guard "Press F12"   -> press('f12')
 |- guard "Press Enter" -> press('enter')
 |- guard "Press Esc"   -> press('esc')
\end{lstlisting}

\begin{lstlisting}[style=jsonsmall]
{
  "type": "object",
  "required": ["reason", "cover_ms", "root_id", "nodes"],
  "properties": {
    "reason":   {"type": "string", "minLength": 1},
    "cover_ms": {"type": "integer", "minimum": 1},
    "root_id":  {"type": "string", "minLength": 1},
    "nodes": {
      "type": "array", "minItems": 1, "maxItems": 10,
      "items": {
        "type": "object",
        "required": ["id", "depth", "guard", "action", "confidence",
                     "deadline_ms", "risk", "expected_state", "children"],
        "properties": {
          "id":             {"type": "string", "minLength": 1},
          "depth":          {"type": "integer", "minimum": 0, "maximum": 2},
          "guard":          {"type": "string", "minLength": 1},
          "action":         {"type": "string", "minLength": 1},
          "confidence":     {"type": "number", "minimum": 0, "maximum": 1},
          "deadline_ms":    {"type": "integer", "minimum": 1},
          "risk":           {"enum": ["low", "medium", "high"]},
          "expected_state": {"type": "string", "minLength": 1},
          "children": {
            "type": "array", "maxItems": 3,
            "items": {"type": "string", "minLength": 1}
          }
        },
        "additionalProperties": false
      }
    }
  },
  "additionalProperties": false
}
\end{lstlisting}

Validation beyond the grammar: executable-action grammar and allowlist,
\texttt{cover\_ms} against the measured planner budget, guard presence on
every executable branch, child-id resolution and reachability, and
node/depth/branch budgets. (The fallback path is a runtime property --
observer abstention -- not a validated tree feature; see
Section~\ref{sec:method-runtime}.) The grammar
backend does not enforce \texttt{maxItems}, which is exactly the gap the
validator plus bounded retry closes (Section~\ref{sec:method-planner}).

\section{Observer prompt and a routed example}
\label{app:observer}

The complete fast-observer prompt template (\texttt{fast\_observer.txt}),
mirroring the tree-schema listing above:

\begin{lstlisting}
You are a fast visual branch selector. Inspect the
screenshot and the eligible branch guards.

Return exactly one JSON object: {"branch_id":"<id or
NONE>","confidence":0.0,"reason":"brief visual evidence"}.

Select a branch only when its guard is visibly satisfied.
Do not generate actions, future plans, markdown, or extra
text. If the screenshot is ambiguous, unchanged, or matches
no guard, return NONE. Be concise.
\end{lstlisting}

A real routed decision (B.1, seed 63010). Eligible guards passed with the
gated frame (the compiled tree's three children; deadlines in ms):

\begin{lstlisting}
[{"id": "prompt_f12",   "guard": "Prompt requests F12",   "deadline_ms": 500},
 {"id": "prompt_enter", "guard": "Prompt requests Enter", "deadline_ms": 500},
 {"id": "prompt_esc",   "guard": "Prompt requests Esc",   "deadline_ms": 500}]
\end{lstlisting}

Observer response (24 completion tokens, 246.9\,ms):

\begin{lstlisting}
{"branch_id":"prompt_enter","confidence":1.0,
 "reason":"The prompt explicitly requests pressing Enter."}
\end{lstlisting}

The runtime then checked eligibility, the planner-side branch confidence
($0.95 \geq \tau_m = 0.45$), and decision age (249.3\,ms $\leq$ 500\,ms
deadline) before executing the pre-authorized
\texttt{pyautogui.press('enter')}. The preceding static frame had returned
\texttt{NONE} (the abstain path exercised in the same trial).

\section{Per-model serving provenance}
\label{app:provenance}

All models served with vLLM~0.25.0, BF16, pinned Hugging Face revisions,
\texttt{max\_model\_len} 8192, on RTX 6000 Ada GPUs (TP=2 except the 7B at
TP=1). The FlashInfer sampler is disabled by launch environment (its JIT is
incompatible with the installed CUDA headers \emph{and} the env var is part
of the compile-cache key). Thinking modes are disabled where the chat
template honors it; one model required a no-think template override
(pre-filled empty think block), documented as its adaptation.

\begin{table*}[t]
\centering
\caption{Frozen serving identities. The compile-cache key is vLLM's
\texttt{torch.compile} cache identity, verified in the server log before
every data-collecting run; revisions are abbreviated to eight hex digits.}
\label{tab:provenance}
\small
\begin{tabular}{llll}
\toprule
Model & HF revision & Compile-cache key & Launch \\
\midrule
Holo-3.1-35B-A3B & \texttt{2bdb9285} & \texttt{cca9e728f1} & TP=2 \\
Qwen3.6-35B-A3B & \texttt{995ad96e} & \texttt{2e9fc7f7e2} & TP=2 \\
UI-Venus-1.5-30B-A3B & \texttt{b6e8f31f} & \texttt{8fc4b0969f} & TP=2 \\
EvoCUA-32B & \texttt{ce055343} & \texttt{0f5d6516ae} & TP=2, no-think template \\
UI-Ins-32B & \texttt{9ee51926} & \texttt{f1cbc332a7} & TP=2, forced bf16 \\
UI-TARS-1.5-7B & \texttt{683d002d} & \texttt{ff3c9ea5a8} & TP=1 \\
\bottomrule
\end{tabular}
\end{table*}

\paragraph{Reproducibility note (serving numerics).}
The compile-cache key is part of the experimental condition. In the planner
diagnostic stage, adding \texttt{--revision} and an offline flag to an
otherwise identical launch changed the cache key, forcing fresh kernel
autotuning; tree validity fell from 88.1\% to 18.3\% on identical weights,
sampler, prompts, and pixel-identical screenshots, and recovered exactly
when the original byte-identical launch (and thus the original compiled
kernels) was restored. The failure flavor was a single token-level mode
flip (the root node's \texttt{"action": "WAIT"} field present vs.\
omitted). Consequences adopted study-wide: byte-exact launch lines are part
of provenance; the cache key is read from the server log and verified
before data collection; per-model keys are frozen (Table~\ref{tab:provenance});
and any key change voids collected data.

\paragraph{Token-normalized costs.}
Each model's decode rate is fit from a forced-length token bench in both
prompt regimes (planner-sized and observer-sized inputs;
$R^2 \geq 0.99$), giving tok/s and a fixed per-call overhead. Critical-path
costs in the paper convert measured wall-clock latencies through the
serving-stack rate (e.g., 181 tok/s for the primary model) into
output-token equivalents, so a reader can rescale every window and latency
to their own hardware.

\section{Round-1 addition: five-outcome event family (EXP-A)}
\label{app:round1-expA}

To probe the coverage rule of Section~\ref{sec:method-branching} beyond the
degenerate three-outcome case, we added a five-outcome benchmark scenario
(uniform over F12/Enter/Esc/F2/F9, seed-derived and byte-stable against the
historical three-key schedule) and pre-registered endpoints, seeds, and
predictions before data collection.

\paragraph{Validity gate (planner-side).} With $K_{\max}=5$ and no prompt
adaptation, tree validity was 29/30 and 29/30 trees covered all five
outcomes: the planner ports to the larger outcome family unchanged.

\paragraph{Window calibration and a null primary result.} The
pre-registered contested-window scan \{500--800\,ms\} floored R0 (the live
reactive baseline) at 0/10 on the five-outcome event, while a concurrent
sanity check on the three-key event reproduced historical behavior --- the
larger outcome set alone slows the reactive model's decisive call by
$\approx$150--200\,ms, shifting the contested band right. Under the
pre-declared widen-once rule we scanned \{850--1000\,ms\}, declared
900\,ms (scan R0 $=0.50$) before any treatment trial, and ran the primary
42-pair block. The result is null: R0 went 42/42 (ceiling) while T2 went
38/42 (0.905); exact McNemar $p=0.125$ with all four discordant pairs
favoring R0. The mechanism is serving non-stationarity: R0's reaction
distribution shifted $\approx$100\,ms faster between scan and block (block
p50/p95 $= 730/810.6$\,ms) on the same server process and verified frozen
compile-cache key, so the window that was contested at scan time had
ceilinged by block time. We report the null verbatim rather than moving the
window post hoc; it is itself evidence for the paper's core caution that
$\sim$100-ms-scale serving drift changes outcomes at contested windows.
Planner-side gates at 900\,ms were unaffected: validity 0.952, outcome
recall 0.929, visible-guard branch accuracy 0.974, zero incorrect actions
from T2.

\paragraph{Exploratory contested block.} Because the declared window
ceilinged, we ran one additional 42-pair block at 750\,ms -- chosen from
the primary block's measured R0 reaction distribution (p50 $=730$\,ms),
declared before any 750\,ms trial, and labeled exploratory since its window
was not set by the pre-registered scan procedure. There the regime is
contested and the paper's mechanism reproduces on the five-outcome family:
T2 $=25/42$ (0.595) vs.\ R0 $=11/42$ (0.262), discordant pairs 16/2 in
T2's favor, exact McNemar $p=1.3\times10^{-3}$. T2's decline from 0.905 at
900\,ms is late-routing pressure (observer decode $\approx$283\,ms against
the tighter window), the same dominant miss mode as the main-benchmark
taxonomy (Section~\ref{sec:results}).

\paragraph{Branch-budget ablation: the knee moves to $k=5$.} The
pre-registered prediction was that the knee of
Section~\ref{sec:ablation-kn} tracks the outcome-set size. It does
(Table~\ref{tab:expA-kcells}): T2 success peaks at $K_{\max}=5$ and is
degraded on \emph{both} sides, so the optimum is interior, not a
saturation.

\begin{table}[t]
\centering
\caption{Five-outcome family, branch-budget cells (42 pairs each,
900\,ms). The pre-registered knee at $k=$ outcome-set size $=5$ is
confirmed; the $k{=}2$ validity valley of Section~\ref{sec:ablation-kn}
reappears at $k{=}3$.}
\label{tab:expA-kcells}
\small
\begin{tabular}{lccl}
\toprule
$K_{\max}$ & Validity & T2 success & Dominant failure \\
\midrule
1 & 24/42 & 2/42 (0.048) & schema, misroute \\
2 & 37/42 & 12/42 (0.286) & misroute, routing \\
3 & \textbf{2/42} & 1/42 (0.024) & budget refusal \\
5 & 33/42 & \textbf{32/42 (0.762)} & schema \\
7 & 39/42 & 19/42 (0.452) & armed catch-all \\
\bottomrule
\end{tabular}
\end{table}

The $k{=}3$ valley has an identified mechanism we call \emph{budget
refusal}: 38/42 failures are the post-hoc validator rejecting
\texttt{children} for \texttt{maxItems:\,3} because the planner, able to
see all five prompt outcomes, emits all five children rather than
truncating to its budget (the bounded retry then re-fails). This
generalizes the $k{=}2$ uncanny valley observed on the three-outcome
family (validity 0.405 there, 0.048 here): the valley sits at budgets just
below the visible outcome-set size, where the planner will neither
specialize (as at $k \leq 2$) nor cover (as at $k \geq 5$).

\paragraph{Mis-set budgets execute wrong actions.} The budget cells bound
the zero-incorrect-actions property of Section~\ref{sec:method-runtime}:
wrong keys were actually pressed in 13/42 trials at $k{=}1$, 12/42 at
$k{=}2$, and 15/42 at $k{=}7$ --- versus 0/42 at the matched budget
$k{=}5$. The two mechanisms are instructive. Under-budgeted planners
($k<5$) write \emph{generic} guards (``Prompt visible'' rather than
``Prompt requests F12''), which the observer matches on any outcome,
firing the covered key. The over-budgeted planner ($k{=}7$) fills its
surplus slots with catch-all branches (``Unknown prompt'') that
\emph{carry a default action} (28/39 valid trees); the observer correctly
routes ambiguous frames to the catch-all, and the default action fires.
Guard specificity and a passive fallback are planner behaviors, not
validator-enforced invariants, so the branch budget is a correctness
parameter: it should be set to the enumerable outcome-set size, and
mis-setting it in either direction converts misses into wrong actions.

\paragraph{Uneven outcome probabilities (coverage rule, Eq.~3).} A second
arm reweights the five outcomes to $.35/.30/.20/.10/.05$ under a
$K_{\max}=4$ budget, so covering to $\rho=0.95$ requires dropping exactly
the rarest outcome (F9). The pre-registered planner-side prediction held:
30/30 trees valid, and 29/30 dropped exactly F9, keeping the four most
likely branches; in the paired block (42 pairs, 900\,ms) 39/42 compiled
trees again dropped exactly F9.

The paired endpoint, reported verbatim, was zero: T2 $0/42$ against R0
$39/42$. The traces attribute all of it to a single design interaction we
had not anticipated: told the outcome distribution, the planner copies the
\emph{priors} into each branch's \texttt{confidence} field (leaf
confidences exactly $.35/.30/.20/.10$ across the block, versus
guard-conditional $0.80$--$0.95$ in the even family), and the runtime's
suppression check~(ii) -- the planner-confidence floor $\tau_m=0.45$ --
then rejects every route. All 80 observer calls that selected a branch
were suppressed as \texttt{low\_confidence}, including plainly correct
ones; replaying the rejected routes against ground truth shows 40/42
would have fired the right key (0 the wrong key) with decision ages
$\leq 394$\,ms against a 500\,ms branch deadline. The observer and the
drop-rarest trees are sound; the floor alone zeroes the arm. The lesson
mirrors the budget finding: $Q_v$ does double duty as calibrated action
confidence and, once priors are visible to the planner, as an outcome
prior -- and $P(\text{outcome})$ is not
$P(\text{action correct} \mid \text{guard matched})$. Any outcome family
whose maximum prior sits below $\tau_m$ is unroutable by construction.
Separating the two fields (a branch \texttt{prior} distinct from
\texttt{confidence}), or gating outcome-branch nodes on observer-match
confidence instead, is the recorded fix direction; we report the null
rather than patching and rerunning.

\section{Round-1 addition: prep-budget sweep and delay jitter (EXP-B)}
\label{app:round1-expB}

\paragraph{What ``prep'' means operationally.} Prep is the budget from task
start (quiet screen visible) to event onset -- the time the planner has to
compile, validate, and arm a tree. The sweep uses a \emph{prep-race} mode
that arms the event countdown \emph{before} planning starts, so the
declared delay is the whole prep budget; in all other experiments the
countdown is armed only after the tree is ready. The window was declared
from a fresh reactive-only scan (700\,ms, R0 $=0.50$), and the planner's
latency was stable across cells (p50 $\approx$ 1.8\,s, p95 $\approx$
1.9\,s), so Eq.~(2) predicts recovery at
$L_{p95} + M \approx 2.4$\,s.

\paragraph{The crossover brackets $L_{p95}+M$, and starved prep fails
safe.} With prep $\in \{0.5, 1\}$\,s -- below the planner's own p50 -- T2
scores $0/42$ in both cells: the event fires mid-compile, no tree is armed,
and every failure is a \emph{miss}, never a wrong action (0 wrong
executions across all 168 sweep T2 trials; the dominant failure stage is
the change gate firing with no tree to route into). At prep $=2$\,s, just
below the predicted threshold, T2 recovers to $36/42$ but shows no
advantage over the reactive arm ($37/42$, discordant $0/1$, $p=1$): the
arming margin is thin and six trials miss at the benchmark deadline. At
prep $=4$\,s, above the threshold, T2 lands $42/42$. The deployer-facing
summary: AAPT needs its own measured $L_{p95}+M$ of quiet screen before
the event; give it less and it degrades to the reactive baseline or below,
but it degrades by \emph{missing}, not by acting wrongly.

One caveat is reported verbatim: the reactive arm scored $0/42$ in the
4\,s cell, and the trial payloads attribute this to \emph{format drift},
not timing -- after $\sim$8 consecutive idle polls the reactive model
abandons the executable-string action format for a nested object that the
executor rejects. T2's $42/42$ recovery stands on its own; the $42/0$
discordant split against R0 in that cell partially reflects this unrelated
reactive fragility (long idle phases degrade reactive format adherence),
which we flag as its own finding about reactive baselines rather than
absorb into the AAPT comparison.

\paragraph{Delay jitter: the fixed-delay leak concern, answered with
data.} A paired block with the event delay drawn per seed from
$U(500, 1000)$\,ms (instead of the fixed 750\,ms) leaves both arms'
behavior consistent with the mechanism claim: T2 holds $36/42$ (Fisher vs.\
a fixed-delay reference block at the same window: $p=0.11$, n.s.), and
within the jitter block T2 beats R0 at $15/0$ discordant pairs
($p=6.1\times10^{-5}$). Nothing in the runtime conditions on the delay
value -- routing is change-gate plus observer driven -- and randomizing the
delay does not remove the gain, so the fixed 750\,ms delay of the main
protocol is not an information leak. The reactive arm's cross-block Fisher
is significant ($21/42$ jittered vs.\ $38/40$ fixed, $p=4.2\times10^{-6}$),
but the attribution is confounded: the same fixed-delay window swung from
R0 $=0.50$ (scan draw) to $0.95$ (reference draw) across seed draws, and
the prep sweep shows R0's decisive-call arrival is phase-locked to the
delay value (success $37 \to 26 \to 37$ across fixed prep values). Both
effects sit on the reactive side; the direction -- randomization makes the
\emph{baseline} no better and leaves the treatment unchanged -- is the
opposite of what a timing leak in AAPT's favor would produce.

\section{Round-1 addition: oracle-routing probe on UI-Venus (EXP-D)}
\label{app:round1-expD}

To test whether the routing gate~(c) of Table~\ref{tab:crossmodel} is
causal rather than correlational, a pre-registered probe re-ran the
UI-Venus-1.5-30B-A3B pairing with the live observer replaced by
ground-truth routing (condition O1: zero-latency keystroke oracle, gated
on the same change signal, able to fire only branches present in the
compiled tree). Everything else was frozen -- same one-round-adapted
prompts, same 600\,ms window as the original paired block, fresh seeds,
serving compile key verified unchanged. Our committed prediction was that
oracle routing would \emph{not} beat R0, on the premise that the original
block's tree coverage (outcome recall 0.22) caps the oracle below R0.

The prediction failed: O1 succeeded on 22/42 vs.\ R0 12/42 (discordant
13/3, exact McNemar $p=0.021$), while R0 reproduced its original rate
(12/42 vs.\ 13/42). The decomposition is exact, with no residual: all 22
successes are precisely the trials whose tree was schema-valid \emph{and}
covered the true outcome; the oracle abstained on all 10 valid-but-
uncovered trees (deadline miss, no wrong action) and 10 trials failed
schema validation. O1 executed zero incorrect actions; R0 executed wrong
actions in 30/42 trials. The prediction failed because its premise did:
this session's unseeded planner samples had recall 0.59 (vs.\ 0.22 in the
original block, identical config and compile key -- a between-session
distribution shift on a planner already documented as numerically
knife-edged, Section~\ref{sec:setup-serving}). The licensed reading is
therefore conditional and two-sided: routing quality is causally binding
(perfect routing alone turns the tie into a significant win), and planner
outcome-recall remains co-binding (the oracle lands exactly at its
coverage ceiling, $22/42 = $ the covered fraction). The UI-Venus row of
Table~\ref{tab:crossmodel} is annotated accordingly.

\section{Round-1 addition: survey of documented response windows (EXP-E)}
\label{app:round1-expE}

To situate the benchmark's contested band in deployed systems rather than
constructed ones, we surveyed documented response windows -- deadlines a
user or agent must act within (\emph{react}) or transient UI that is only
visible, and hence only actionable, for a bounded time (\emph{display}).
Figure~\ref{fig:deadline-survey} (main text) places the 19 documented
windows of Table~\ref{tab:deadline-survey} on a log axis against the
paper's contested band and the measured reactive p50.

\begin{table*}[t]
\centering
\caption{Survey of documented response windows (EXP-E). Sources are
primary documentation; ``configurable'' notes whether deployments can
lengthen the window.}
\label{tab:deadline-survey}
\scriptsize
\setlength{\tabcolsep}{4pt}
\begin{tabular}{lllll}
\toprule
System & Window & Config. & Class & Source \\
\midrule
BIOS ultra-fast-boot F2/DEL & $\approx$0--200\,ms & fast-boot toggle & react & vendor fast-boot docs \\
SF3 3rd Strike parry & 83--167\,ms & no & react & wiki.supercombo.gg (5--10 frames @60\,fps) \\
Android double-tap timeout & 300\,ms & no & react & developer.android.com \texttt{ViewConfiguration} \\
Windows double-click default & 500\,ms & yes (max 5\,s) & react & learn.microsoft.com \texttt{GetDoubleClickTime} \\
Android long-press timeout & 500\,ms & accessibility & react & developer.android.com \texttt{ViewConfiguration} \\
Android \texttt{Toast.LENGTH\_SHORT} & 2\,s & no & display & Android docs constant \\
Android \texttt{Toast.LENGTH\_LONG} & 3.5\,s & no & display & Android docs constant \\
Flutter SnackBar default & 4\,s & yes & display & api.flutter.dev \texttt{SnackBar.duration} \\
MUTCD WALK interval & 4--7\,s & agency-set & react & mutcd.fhwa.dot.gov \S4I.06 \\
Material 3 snackbar guidance & 4--10\,s & yes & display & m3.material.io snackbar spec \\
Bootstrap 5 toast default & 5\,s & yes & display & getbootstrap.com toasts \\
Windows 11 notification banner & 5\,s & min 5\,s & display & learn.microsoft.com \\
GRUB menu timeout default & 5\,s & yes & react & GNU GRUB manual \texttt{GRUB\_TIMEOUT} \\
iOS temporary notification banner & 5--7\,s & no & display & discussions.apple.com \\
ALKS takeover request & 10\,s & regulated & react & UNECE R157 transition demand \\
WCAG 2.2.1 extend-session response & 20\,s & regulated floor & react & w3.org WCAG SC 2.2.1 \\
TOTP time step & 30\,s & rarely & react & RFC 6238 \\
Duo Push approval timeout & 60\,s & no & react & help.duo.com \\
OpenSSH \texttt{LoginGraceTime} default & 120\,s & yes & react & \texttt{sshd\_config} \\
\bottomrule
\end{tabular}
\end{table*}

The population divides at the measured reactive latency. Below and inside
the contested band -- parry windows, double-tap/double-click/long-press
timeouts, fast-boot hotkey windows -- a 567--730\,ms reactive loop is
structurally excluded or coin-flip, and pre-computation is the only
mechanism that acts in time. In the 1--10\,s band (toasts, snackbars,
banners, GRUB, WALK intervals) a reactive VLM loop competes but with thin
margins under load or multi-step responses, and the $\pm$100\,ms serving
non-stationarity we measured (Appendix~\ref{app:round1-expA}) is a
meaningful fraction of the shortest of these. At or above 10\,s
(regulated takeover, WCAG, TOTP, push approval, SSH) the reactive loop
suffices and AAPT's premium is not justified. The survey bounds the claim
honestly in both directions: real sub-second windows exist in deployed
systems, and most windows above a few seconds do not need anticipation.

\section{Method details moved from the main text}
\label{app:method-details}

\subsection{Branch allocation rule}

The generation-time allocation rule for the per-node branching factor
$K_v$ of Section~\ref{sec:method-branching} is
\begin{equation}
K_v =
\begin{cases}
1, & \text{confidence} \geq \tau_c \text{ and risk} \leq \tau_r,\\
\min(K_{\max}, m_v), & \text{outcome uncertain},\\
0, & \text{safe fallback / replan required},
\end{cases}
\end{equation}
subject to global budgets: per-node limit $K_{\max}$, maximum depth
$n_{\max}$, total node budget, coverage target $T_{\mathrm{cover}}$, and
risk-specific open-loop horizons. The hard budgets ($K_{\max}$,
$n_{\max}$, node count, \texttt{cover\_ms}) are enforced numerically by
the post-hoc validator, whereas $\tau_c$, $\tau_r$, $\rho$, and $m_v$ are
\emph{design rules realized as planner-prompt instructions}, not
runtime-enforced thresholds -- the planner is asked to branch on uncertain
outcomes and cover the likely ones, and the rule describes the behavior
this induces.

\subsection{Engineering findings behind guided-decoding planning}

Both findings were initially misdiagnosed as capability gaps.
\textbf{Schema shape dominates grammar constraints.} With the original
schema, which expressed ``an action or an action array'' as a JSON-Schema
\texttt{oneOf} alternation, tree validity was 88.1\% (238/270) and the
dominant failure was a node omitting its required action field entirely.
Removing the alternation (scalar \texttt{action} only) fixed validity to
60/60 \emph{even without any decoding constraint}, while
grammar-constrained decoding with the original schema reached only 50/60.
The deficit was induced by the schema, not by the model; guided decoding
plus the simplified schema yields 230/231 (99.57\%) pooled validity, and a
planner LoRA fine-tune that had been provisionally supported by the
earlier numbers was cancelled.
\textbf{Post-hoc validation plus bounded retry closes the residual gap.}
The grammar backend does not enforce JSON-Schema array-length bounds
(\texttt{maxItems}), so a rare over-wide or unreachable tree still
decodes; a strict validator (action grammar and allowlist, deadline and
coverage checks, guard presence on every executable branch, reachability,
budget enforcement) rejects it, and the planner resamples once at
unchanged temperature. In live operation the retry fired on 7/336 trials
and recovered 4, leaving 0.89\% final invalidity.

\subsection{Change-gate constants and latency accounting}

The pixel-difference gate that feeds the observer is fully specified by
three constants, fixed across all experiments: a frame is ``changed'' when
at least 0.5\% of pixels (\texttt{min\_changed\_fraction} $=0.005$) differ
from the previous gated frame by $\geq 20$ grayscale levels
(\texttt{pixel\_threshold} $=20$, absolute difference on grayscale), and a
static-scene probe frame is forced through every 500\,ms
(\texttt{static\_probe\_ms}) so a stalled screen cannot silence the
observer indefinitely. Capture runs at 30\,FPS, so the gate contributes at
most one frame interval ($\leq 33$\,ms, expected $\approx 17$\,ms) between
event onset and the frame the observer decodes; T2's reported reaction
time includes this capture quantization plus observer decode and routing
checks.

\subsection{Model inventory}
\label{app:models}

\begin{table*}[t]
\centering
\caption{Models studied (CUA = computer-use agent; MoE = mixture-of-experts).
Decode is measured observer-regime speed on our serving stack (1--2 RTX
6000 Ada GPUs, tensor parallel 2 for the 30--35B models); ``observer
call'' is the latency of one routing call. MoE models decode roughly
$7\times$ faster than the dense 32B models, one of three gates for the
AAPT gain.}
\label{tab:models}
\small
\setlength{\tabcolsep}{4pt}
\begin{tabular}{llrrl}
\toprule
Model & Class & Decode (tok/s) & Observer call & Role in this paper \\
\midrule
Holo-3.1-35B-A3B & Qwen3.5-MoE, CUA-tuned & $\sim$181 & $\sim$0.2\,s & primary model \\
Qwen3.6-35B-A3B & Qwen3.5-MoE, generalist & 180--201 & $\sim$0.15--0.45\,s & confirmation model \\
UI-Venus-1.5-30B-A3B & Qwen3VL-MoE, CUA-tuned & $\sim$170 & $\sim$0.15\,s & gate (c) boundary \\
EvoCUA-32B & Qwen3VL dense, CUA-tuned & $\sim$25 & $\sim$2.5\,s & gate (a) boundary \\
UI-Ins-32B & Qwen2.5-VL dense, CUA-tuned & $\sim$26 & $\sim$1.5\,s & gates (a)+(b) boundary \\
UI-TARS-1.5-7B & Qwen2.5-VL dense, CUA-tuned & $\sim$59 & $\sim$0.5--1.1\,s & gate (b) boundary \\
\bottomrule
\end{tabular}
\end{table*}

Table~\ref{tab:models} lists the six checkpoints with measured
observer-regime decode speeds and their roles in the study:
\holo{}~\citep{holo2026}, \qwen{}~\citep{qwen36},
UI-Venus-1.5-30B-A3B~\citep{uivenus}, EvoCUA-32B~\citep{evocua},
UI-Ins-32B~\citep{uiins}, and UI-TARS-1.5-7B~\citep{uitars}.

\subsection{Deliberately unclaimed extensions}

Two natural extensions from the original design are not claimed:
asynchronous tree refresh (overlapping planning windows, so a successor
tree is compiled while the current one controls the fast path) and
reusable policy memory (promotion of repeatedly validated subtrees into
PreAct-style procedural skills, with promotion gates on repeated success
and revocation support). Both remain design sketches.

\section{Sizing-rule details: grids, frontier, and token sweep}
\label{app:sizing}

\subsection{The $k \times n$ frontier}

Six cells ($k \in \{1,2,3\}$, $n \in \{1,2\}$, per-cell generated
guided-decoding schemas), 42 pairs per cell against a shared reactive arm,
paired by seed. The pre-registered 650\,ms primary sat at R0 ceiling this
session (0.952; serving was faster than in the confirmatory stage), so the
window was recalibrated by the fixed reactive-only protocol to 600\,ms (R0
$=0.548$ on the full 42 primary seeds) -- already a pre-registered
secondary -- with 650\,ms retained as a ceiling anchor and 575\,ms as a
harder secondary. Costs are reported in milliseconds and in output-token
equivalents at the measured 181 tok/s.

\begin{table*}[t]
\centering
\caption{The $k \times n$ frontier at 600\,ms (shared R0 $=0.548$). The
node budget is the schema's maximum tree size; critical-path cost is the
observer path in output-token equivalents (p50). Both $k{=}3$ cells
dominate R0 (19 T2-only vs.\ 0 R0-only pairs); both are 42/42, but $n{=}2$
spends a $3.25\times$ node budget on an identical realized tree.}
\label{tab:frontier}
\small
\begin{tabular}{lrrrllcr}
\toprule
Cell & $k$ & $n$ & Budget & T2 succ.\ $\uparrow$ & Validity $\uparrow$ & T2-/R0-only & Crit.\ path (tok) $\downarrow$ \\
\midrule
$k1\,n1$ & 1 & 1 & 2 & 6/42 (0.143) & 36/42 (0.857) & 4 / 21 & 51.9 \\
$k1\,n2$ & 1 & 2 & 3 & 1/42 (0.024) & 5/42 (0.119) & 1 / 23 & 54.1 \\
$k2\,n1$ & 2 & 1 & 3 & 8/42 (0.190) & 17/42 (0.405) & 2 / 17 & 54.5 \\
$k2\,n2$ & 2 & 2 & 7 & 6/42 (0.143) & 12/42 (0.286) & 5 / 22 & 54.7 \\
$\mathbf{k3\,n1}$ & 3 & 1 & 4 & \textbf{42/42 (1.000)} & \textbf{42/42 (1.000)} & \textbf{19 / 0} & 58.8 \\
$k3\,n2$ & 3 & 2 & 13 & 42/42 (1.000) & 42/42 (1.000) & 19 / 0 & 56.9 \\
\bottomrule
\end{tabular}
\end{table*}

The ordering is fully window-robust: at the harder 575\,ms secondary (R0
$=0.071$), $k3\,n1$ still lands 42/42 while all $k<3$ cells stay floored.
$k{=}1$ stays more valid (0.857) than $k{=}2$ (0.405) but commits to a
single outcome and is coverage-floored. On depth: the task has one
decision step, and the planner builds the same four-node depth-1 tree even
when 13 nodes and depth 2 are allowed (realized tree p50 $=4$ in both
$k{=}3$ cells); for $k<3$ extra depth actively collapses validity further
($k1\,n2$: 0.119). The critical-path cost is nearly flat across all six
cells (observer p50 287--325\,ms $\approx$ 52--59 token-equivalents):
accuracy is set by planner-side validity and coverage budget, not observer
latency. Planner prep cost (off the critical path) grows from
$\sim$1.0\,s to $\sim$1.8\,s with budget.

\subsection{Observer token-cap sweep}

\begin{table}[t]
\centering
\caption{Observer token-cap sweep, pooled over the two contested windows
(eight cells, 42 pairs each).}
\label{tab:observer-tokens}
\small
\begin{tabular}{lrrl}
\toprule
Cap & Pooled T2 & Rate & Note \\
\midrule
32 & 48/84 & 0.571 & truncates JSON (27 fail.) \\
\textbf{64} & \textbf{70/84} & \textbf{0.833} & winner: 0 fail., lowest p95 \\
96 & 69/84 & 0.821 & statistical tie with 64 \\
128 & 58/84 & 0.690 & ineligible: branch acc.\ 0.829 \\
\bottomrule
\end{tabular}
\end{table}

The failure modes bracket the knee (Table~\ref{tab:observer-tokens}): 32
tokens truncates the observer's JSON mid-object (pooled 0.571, 27
failures), while 128 tokens adds latency and drifts branch accuracy below
the gate (0.829). Adopting the 64-token observer lifted the
contested-regime T2 ceiling from $\sim$0.70 (split between wrong-branch
routing and late second observer calls) to 0.79/0.88 at 600/650\,ms, with
zero late routings in all eight cells of the adopting run. The planner
side was hardened in the same stage by the bounded retry-on-invalid:
acceptance 30/30 valid, and over the 336 live trials the retry fired
seven times and recovered four invalid trees, cutting final planner
invalidity from 2.08\% to 0.89\%.

\subsection{Window sensitivity and the recalibration protocol}
\label{app:window-sensitivity}

The contested window is a property of the model-serving-seed triple, not
a benchmark constant: it moved from 600--650\,ms (shared-server sessions)
down to $\sim$600\,ms and below (dedicated server, faster decode), and
across seed draws it can vanish entirely into a near-vertical cliff. Two
protocol rules kept this honest: (i) windows are recalibrated only from
reactive-only scans before any treatment data, and only toward windows
that were pre-registered as secondaries; (ii) when a fresh seed set has
no contested midpoint, the mechanism is tested at a sub-cliff window
instead, where the paired design still discriminates (the reactive arm
cannot reach the deadline that the $\sim$300\,ms observer can). Per-seed
reactive latency is stable across sessions (p50 567\,ms, reproduced
exactly an hour apart), so these shifts are difficulty variance across
42-seed draws, not server drift.

\section{Additional results detail}
\label{app:results-detail}

\subsection{Prediction ledger}
\label{app:ledger-r1}

Table~\ref{tab:ladder} is the paper's prediction ledger: every claim in
the main-text results was declared before its data, including two that
failed, so no result reads as post-hoc selection. Table~\ref{tab:ledger-r1}
adds the entries from the Round-1 revision cycle.

\begin{table*}[t]
\centering
\caption{The prediction ledger: what was declared before each block's
data, and what happened. Failed predictions are retained at full weight,
and every deviation is reported in the listed section.}
\label{tab:ladder}
\scriptsize
\setlength{\tabcolsep}{4pt}
\begin{tabular}{@{}l>{\raggedright\arraybackslash}p{5.6cm}>{\raggedright\arraybackslash}p{8.2cm}@{}}
\toprule
Stage & Declared before data (primary endpoint) & Outcome \\
\midrule
Contested scan & windows, 700\,ms primary (T2$>$R0) & n.s.\ (R0 near ceiling); secondaries carry it (\S\ref{sec:results-headline}) \\
Confirmation & 650\,ms declared primary (T2$>$R0) & \textbf{0.79 vs.\ 0.50, $p=1.8\times10^{-3}$} (\S\ref{sec:results-headline}) \\
Mechanism & conditions, windows (T2 vs.\ R1/P1/O1) & R1$=$P1$=$0; O1 headroom estimate (\S\ref{sec:results-mechanism}) \\
$k \times n$ ablation & grid, 600\,ms primary (frontier shape) & knee at $k{=}$ outcome-set size, $n{=}1$ (\S\ref{sec:ablation}) \\
Fresh-seed repeats & 550\,ms, seeds (direction repeats) & pooled $p=1.8\times10^{-12}$, 40/0 discordant (\S\ref{sec:results-replication}) \\
Boundary map & gate predictions (per-model gates) & each model fails one gate (\S\ref{sec:results-boundary}) \\
Gate confirmation & predictions committed pre-token (2 unseen models) & both graded correct; $p=4.9\times10^{-13}$ (\S\ref{sec:results-replication}) \\
Oracle probe (R1 rev.) & committed: oracle does not beat R0 on UI-Venus & \textbf{prediction failed}: 22/42 vs.\ 12/42, $p=0.021$; routing gate causal (App.~\ref{app:round1-expD}) \\
DynaCU power run (R1 rev.) & 3 episodes, majority-of-3; 80\% concentration rule & confirmed: 100\%/100\% concentration; aggregate stays a tie (\S\ref{sec:results-transfer}) \\
\bottomrule
\end{tabular}
\end{table*}

\begin{table*}[t]
\centering
\caption{Continuation of the prediction ledger: entries
added in the Round-1 revision cycle. As in the main table, every entry
was declared before its block's data and failed predictions are retained
at full weight.}
\label{tab:ledger-r1}
\small
\setlength{\tabcolsep}{4pt}
\begin{tabular}{@{}l>{\raggedright\arraybackslash}p{5.6cm}>{\raggedright\arraybackslash}p{8.2cm}@{}}
\toprule
Stage & Declared before data (primary endpoint) & Outcome \\
\midrule
Five-outcome family & addendum: seeds, windows (T2$>$R0 @900\,ms) & null (R0 ceilinged post-scan); exploratory 750\,ms $p=1.3\times10^{-3}$ (App.~\ref{app:round1-expA}) \\
Budget cells & $k\in\{1,2,3,5,7\}$, knee-at-5 (T2 vs.\ $k$) & confirmed: knee at $k{=}5$, valley at $k{=}3$ (App.~\ref{app:round1-expA}) \\
Uneven arm & drop-rarest ($\rho{=}0.95$, $K_{\max}{=}4$) & 29/30 + 39/42 dropped exactly F9; paired arm nulled by $\tau_m$ prior collision (App.~\ref{app:round1-expA}) \\
Prep sweep & crossover at $L_{p95}{+}M \approx 2.4$\,s (T2 vs.\ prep) & confirmed: 0/42 at 0.5--1\,s, tie at 2\,s, 42/42 at 4\,s; all failures misses (App.~\ref{app:round1-expB}) \\
Delay jitter & T2 unchanged under $U(500,1000)$\,ms & confirmed: $p=0.11$ n.s.; T2$>$R0 15/0, $p=6.1\times10^{-5}$ (App.~\ref{app:round1-expB}) \\
\bottomrule
\end{tabular}
\end{table*}

Table~\ref{tab:ledger-r1} continues the main paper's prediction ledger
with the Round-1 revision entries.

\subsection{Contested-window scan}
\label{app:scan}

\begin{table}[t]
\centering
\caption{Contested-window scan (42 pairs per window). The pre-registered
primary (700\,ms) is n.s.\ because R0 is near ceiling there; the
secondaries carry the contested-regime result. Incorrect actions: 0
everywhere.}
\label{tab:contested-scan}
\scriptsize
\setlength{\tabcolsep}{2.5pt}
\begin{tabular}{lrrcr}
\toprule
Window & R0 & T2 & Disc.\ (T2/R0) & McNemar $p$ \\
\midrule
600\,ms & 4/42 (0.095) & 29/42 (0.690) & 25 / 0 & $6.0\times10^{-8}$ \\
650\,ms & 12/42 (0.286) & 30/42 (0.714) & 19 / 1 & $4.0\times10^{-5}$ \\
700\,ms (prim.) & 37/42 (0.881) & 40/42 (0.952) & 3 / 0 & 0.25 (n.s.) \\
750\,ms & 39/42 (0.929) & 41/42 (0.976) & 2 / 0 & 0.5 (n.s.) \\
\bottomrule
\end{tabular}
\end{table}

Table~\ref{tab:contested-scan} gives the full four-window scan behind
Test 1's contested-regime localization.

\subsection{Cost accounting and miss taxonomy (declared-primary block)}
\label{app:cost}

\begin{table}[t]
\centering
\caption{Per-trial cost accounting, declared-primary block (B.1, 650\,ms,
42 pairs per condition). Completion tokens are means per trial; wall time
is armed-to-terminal.}
\label{tab:cost}
\small
\begin{tabular}{lrr}
\toprule
 & R0 & T2 \\
\midrule
Planner completion tok & 0 & 283.5 \\
Planner latency p50 (s) & -- & 1.81 \\
Observer/decision completion tok & 105.8 & 59.3 \\
Total completion tok & 105.8 & 342.8 \\
Wall time per trial (s) & 1.65 & 3.32 \\
Trees per successful trigger & -- & 1.27 \\
Successes & 21/42 & 33/42 \\
\bottomrule
\end{tabular}
\end{table}

On the declared-primary block (Table~\ref{tab:cost}), T2 generates
$3.2\times$ R0's completion tokens per trial (342.8 vs.\ 105.8): the tree
compilation (283.5 completion tokens, p50 1.81\,s) is the premium, and the
runtime observer is \emph{cheaper} per call than R0's decisive calls (59.3
vs.\ 105.8). Every trial compiled exactly one tree and 33/42 triggered
successfully (1.27 trees per successful trigger); the premium buys the
off-critical-path position, worth paying exactly when the window is
contested. T2's nine misses on the block: 5 routed-late-executed (the
observer chose the right branch but its decision age, 382--446\,ms, plus
execution landed past the server deadline), 3 wrong-branch routes (all
three chose the F12 branch -- an observer bias -- and all three presses
also landed post-deadline, so they score as misses rather than incorrect
actions), and 1 correctly suppressed late decision. On the same block
every pre-registered secondary window is also individually significant
(600\,ms: 30/42 vs.\ 6/42, $p=1.2\times10^{-7}$; 625\,ms: 30/42 vs.\
2/42, $p=7.5\times10^{-9}$; 675\,ms: 35/42 vs.\ 27/42,
$p=7.8\times10^{-3}$), with tree validity 1.0 (625\,ms: 0.976).

\subsection{Confirmation blocks on the untuned generalist}
\label{app:confirmation}

\begin{figure*}[t]
\centering
\includegraphics[width=0.6\textwidth]{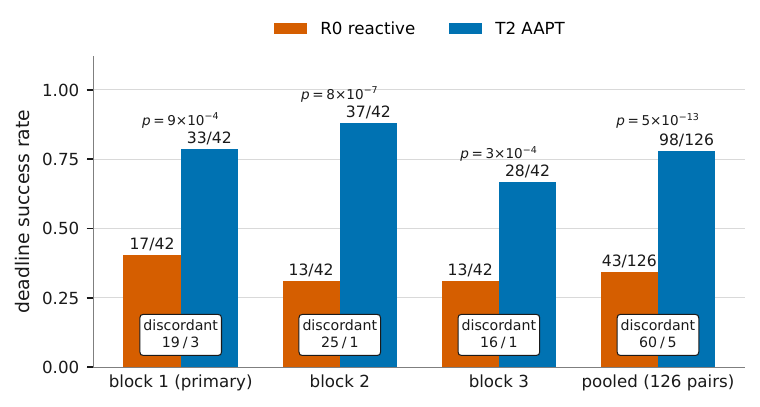}
\caption{Pre-registered confirmation on \qwen{}, an untuned generalist of
the same architecture class as Holo (42 pairs per block, 600\,ms). Boxed
annotations give discordant pairs (T2-only / R0-only), with the exact
McNemar $p$ above each pair. Every block is individually significant and
the direction never reverses.}
\label{fig:f5blocks}
\end{figure*}

Figure~\ref{fig:f5blocks} shows the three 42-pair confirmation blocks of
Section~\ref{sec:results-replication} individually: every block is
significant on its own, the direction never reverses, and the pooled 126
pairs give T2 0.778 vs.\ R0 0.341 (discordant 60/5,
$p=4.9\times10^{-13}$). Gate measurements on this model: decode 180--201
tok/s; tree validity 30/30 with full outcome coverage; branch accuracy
0.969 pooled, expected calibration error 0.020. Incorrect-action rate in
the primary block: R0 0.57, T2 0.00.

\subsection{DynaCU-Bench category table}
\label{app:dynacu}

\begin{table}[t]
\centering
\caption{DynaCU-Bench paired transfer, three-episode power run
(majority-of-3 per task and condition; frozen Holo both arms;
deterministic DOM scoring). Aggregate is a tie ($p=1.0$); the discordant
wins dissociate perfectly by whether the winning response was
pre-enumerable when the tree was compiled (concentration rule
pre-committed: both sides at 100\% vs.\ an 80\% bar).}
\label{tab:dynacu}
\small
\setlength{\tabcolsep}{3pt}
\begin{tabular}{lrrrl}
\toprule
Category & $n$ & Reactive & AAPT & Discordant \\
\midrule
D -- carousels & 10 & 3/10 & 0/10 & 3 react.-only \\
E -- dashboards & 9 & 0/9 & 4/9 & 4 AAPT-only \\
F -- toasts/forms & 10 & 4/10 & 2/10 & 2 react.-only \\
J -- reaction games & 10 & 0/10 & 0/10 & 0 \\
\midrule
Overall & 39 & 7/39 (0.18) & 6/39 (0.15) & 5 / 4, $p=1.0$ \\
\bottomrule
\end{tabular}
\end{table}

Table~\ref{tab:dynacu} gives the per-category outcomes behind the Test-5
dissociation. Episode-level behavior is near-deterministic (5 of 78
task-condition cells non-unanimous). Median decision latency, measured in
the single-episode instrumentation round, was 214\,ms for observer routing
($n=1034$ gated frames) vs.\ 437\,ms per reactive step ($n=345$); the
$\sim$2$\times$ advantage converts into success only when the prepared
action is still valid at event time. No unsafe action occurred in either
arm.

\section{Full limitations register}
\label{app:limitations}

We state every carried caveat from the experimental record in full.

\textbf{Claim scope.} The positive result is bounded twice: it replicates
within the Qwen3.5-MoE architecture class (a computer-use fine-tune and an
untuned generalist), and everywhere else it is capability-gated. Nothing
in this paper supports an architecture-agnostic claim, and three of the
five non-Holo models we attempted show no gain at all.

\textbf{Benchmark breadth.} The main evidence chain (existence,
confirmation, mechanism isolation, replication, ablations) runs on a
single scenario family -- the three-outcome \texttt{key\_prompt} event --
plus one external benchmark. The scenario was designed to isolate the
mechanism, not to represent computer use broadly.

\textbf{Knife-edge contested windows.} Because reaction latency on the
local benchmark is nearly deterministic, whether a 42-seed draw is
contested at a fixed window is almost all-or-nothing: absolute success
rates at a fixed window are not comparable across seed draws (the
benchmark exhibits seed-set difficulty variance, not time drift). Only the
paired per-seed comparisons carry evidential weight, and we caution
against quoting this paper's absolute rates outside their seed sets.

\textbf{Primary-window deviations.} Two pre-registered primary windows
missed their regime: the first contested scan's 700\,ms primary landed at
reactive ceiling and is not significant ($p=0.25$) -- the
contested-regime result at that stage rests on pre-registered secondary
windows and was subsequently confirmed by a fresh declared-primary run at
650\,ms ($p=1.8\times10^{-3}$) -- and the ablation stage's 650\,ms
primary was recalibrated to 600\,ms by the reactive-only protocol. Both
deviations are reported where they occur.

\textbf{A near-miss on a validity sub-target.} The planner-hardening stage
set a pooled-with-history Wilson lower bound of 97\% on tree validity; the
achieved bound is 96.69\% (425/432 pooled). The pre-registered per-run
acceptance (30/30 valid) passed, and we did not expand the sample to chase
the pooled target.

\textbf{Pre-compiled trees cannot bind late parameters.} The
\texttt{click\_target} scenario was not run as a paired comparison because
both arms fail for independent reasons: the reactive arm is floored by the
model's coordinate grounding (0/18 even at relaxed windows; clicks miss a
90\,px target by a 229\,px median despite the model perceiving it), and
the AAPT arm is blocked by construction -- the target's coordinates are
revealed only at fire time, after the tree is compiled, and the observer
returns a branch id, not coordinates. Positional grounding and
observer-parameterized actions are unimplemented extensions.

\textbf{External transfer is a conditional split over small samples.} The
DynaCU aggregate is a tie ($p=1.0$, $n=39$), and the three-episode power
run (majority-of-3) confirms the category dissociation under a
pre-committed concentration rule -- but the split still rests on 4 vs.\ 5
discordant tasks, all AAPT-side wins sit in one category (E dashboards;
the single-episode J wins did not survive the majority rule), and the
result licenses a boundary claim, not a per-category effect size.

\textbf{One boundary model has thin coverage.} The UI-Venus evidence is
two 42-pair blocks -- the original tie (branch accuracy 0.39, outcome
recall 0.22) and the oracle probe that reversed our committed prediction
(Appendix~\ref{app:round1-expD}) -- and the probe also exposed a
between-session shift in the planner's tree coverage (recall 0.22 vs.\
0.59 under identical config and compile key), so per-session tree
statistics for this model should be treated as unstable. Similarly,
UI-TARS's reactive 0/30 arises under the base reactive prompt after the
single permitted adaptation round was spent on its planner; it is a
harness-portability datum, not a capability ceiling for that model.

\textbf{Observer headroom.} The oracle gap
($\mathrm{O1}-\mathrm{T2}\approx0.16$ at the contested windows) means a
sizable fraction of compiled-tree value is lost in live routing; the
reported T2 numbers understate what the trees themselves contain.

\textbf{Unvalidated design components.} Asynchronous tree refresh,
reusable policy memory with promotion gates, and richer simulated
environments are design sketches only
(Appendix~\ref{app:method-details}); no experiment in this paper
validates them. No fine-tuning was performed anywhere: the pre-registered
training triggers (planner and observer LoRA gates) were never met, which
we count as a strength of the frozen-model result but also means this
paper says nothing about what training would add.

\end{document}